\documentclass[runningheads]{llncs}
 
\usepackage[mobile]{eccv}
\usepackage{eccvabbrv}
\usepackage{graphicx}
\usepackage{booktabs}
\usepackage{eso-pic}
\usepackage[accsupp]{axessibility}
\usepackage[pagebackref,breaklinks,colorlinks,citecolor=eccvblue]{hyperref}
\usepackage{multirow}
\usepackage{booktabs}
\usepackage{graphicx}
\usepackage{orcidlink}
\begin{document}
\title{SEE Challenge 2026: Event-Guided Brightness Adjustment Across a Broad Illumination  Range} 
\titlerunning{EBMV 2026 - SEE Challenge}
\author{\small
Yunfan Lu$^{*,1}$, Mingchao Xu$^{*,1}$, 
Hanyu Zhou$^{2}$, Shaoyu Liu$^{3,4}$, Haoyue Liu$^{5}$, Peiqi Duan$^{6}$, Shihan Peng$^{5}$, Yinqiang Zheng$^{7}$, Boxin Shi$^{6}$, Gim Hee Lee$^{2}$, Hui Xiong$^{1}$, Davide Scaramuzza$^{8}$\\~\\
$^1$Hong Kong University of Science and Technology (Guangzhou) $^2$National University of Singapore $^3$Xidian University $^4$Tsinghua University $^5$Huazhong University of Science and Technology $^6$Peking University $^7$The University of Tokyo $^8$Robotics and Perception Group, University of Zurich
}
\authorrunning{Yunfan Lu, Mingchao Xu, et al.}
\institute{\textsuperscript{*} denotes equal contribution}
\maketitle
\AddToShipoutPictureFG*{\AtPageUpperLeft{\raisebox{-8mm}{\makebox[\paperwidth][c]{\footnotesize\color{gray}\shortstack[c]{This paper has been accepted for publication at the\\European Conference on Computer Vision (ECCV) Workshop, 2026}}}}}
\begin{abstract}
Event cameras provide a high dynamic range and preserve brightness-change cues in lighting conditions where conventional RGB frames may be noisy or saturated.
To benchmark event-guided restoration across a broad illumination range, we organized the SEE Challenge 2026 with the Event-Based Multimodal Vision Workshop at ECCV 2026.
The task conditions restoration on one or more RGB frames, synchronized events, and a scalar target-brightness statistic provided by the organizers.
It uses SEE-600K, which contains 610,126 image-event observations from 202 real-world scenes spanning low-light, normal-light, and high-light conditions with illumination variations of up to 1,000$\times$.
The challenge follows an open-system protocol: participants may use different temporal contexts, architectures, pretrained weights, test-time augmentation, and post-processing strategies.
PSNR determines the ranking, and SSIM is reported as a secondary metric.
Around 70 teams registered interest and 15 valid CodaBench submissions were received.
Six distinct teams completed organizer-side identity and technical verification, provided method descriptions, checkpoints, inference code, and instructions, and are included in the verified open-system ranking reported here.
Beyond the ranking, this report analyzes exposure subsets, semantically distinct test cases, a shared failure pattern, system design choices, and inference strategies.
The top systems obtain closely spaced average scores, while the best-performing method varies across cases and metrics; under severe underexposure, all verified systems retain visible local errors.
More details are available at \href{https://www.codabench.org/competitions/16195/}{SEE Challenge 2026}.
\end{abstract}

\section{Introduction}
\label{sec:intro}

Visual systems often encounter illumination changes that exceed the dynamic range of conventional cameras~\cite{koshel2012illumination}.
Mobile robots may move from dim indoor spaces to sunlit entrances, while autonomous vehicles may encounter tunnels, reflective roads, and bright headlights within the same sequence~\cite{desouza2002vision,shariff2024event}.
Under these conditions, dark regions can be dominated by noise and bright regions can become saturated.
Once useful measurements have been severely degraded, recovering faithful structure and color from an RGB image alone becomes difficult.

\begin{figure*}[t!]
    \centering
    \includegraphics[width=0.99\textwidth]{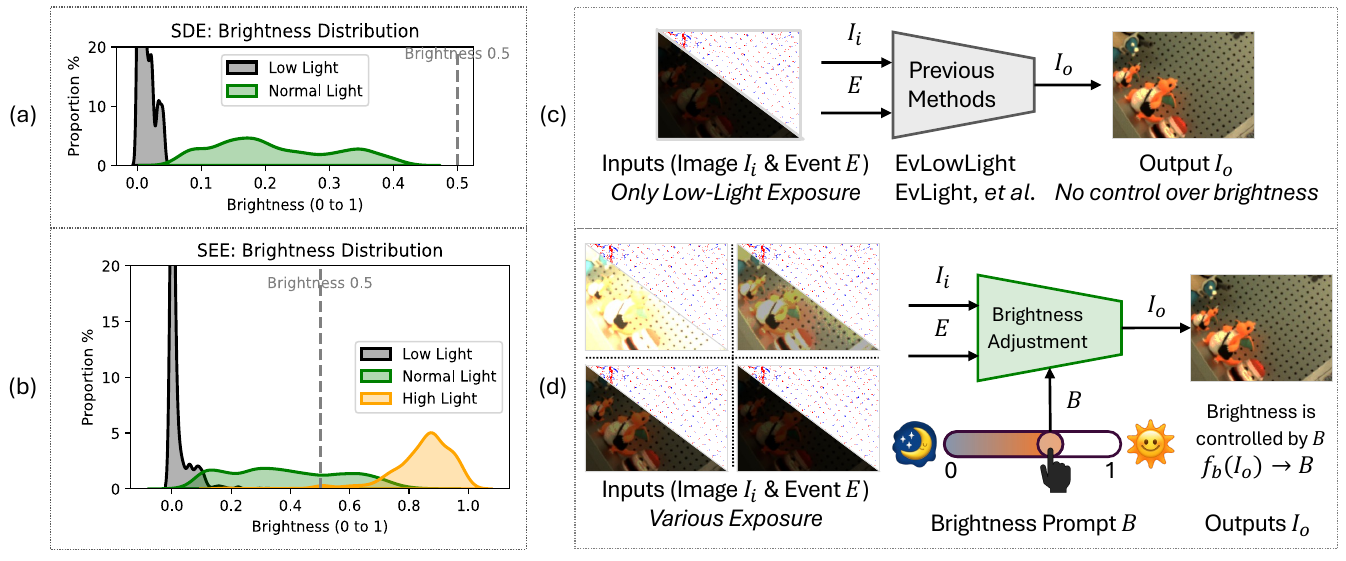}
    \caption{\small Adapted from SEE-Net~\cite{lu2025SEE}. Brightness distributions and task settings of SDE~\cite{chen2025evlight++} and SEE-600K~\cite{lu2025SEE}. (a) SDE covers a low-to-normal brightness range. (b) SEE-600K spans low-light, normal-light, and high-light conditions. (c) Earlier low-light methods map dark inputs to a fixed normal-light target. (d) The official SEE Challenge evaluation conditions restoration on a target-brightness statistic $B$ derived from the reference image.}
    \label{fig:1-SEE-Setting}
\vspace{-15pt}
\end{figure*}

This problem has motivated extensive work in computational photography and image restoration~\cite{guo2020zero,ma2022toward,guo2016lime,xu2023low}.
Multi-exposure high dynamic range imaging combines observations captured at different exposure levels, but remains sensitive to scene motion and alignment~\cite{huo2024multi,kim2021end,tan2023deep}.
Learning-based approaches address low-light enhancement, exposure correction, RAW reconstruction, and image signal processing using learned priors~\cite{wang2018gladnet,afifi2021learning,conde2025raw,archana2024deep}.
Nevertheless, a frame-based method must infer missing information from measurements that may already be clipped or noise-dominated~\cite{gehrig2024low}.

Event cameras asynchronously record per-pixel brightness changes and provide complementary measurements~\cite{chakravarthi2024recent}.
Their high temporal resolution and wide dynamic range can preserve contrast and motion cues when conventional frames are degraded.
Events do not directly provide dense color or absolute brightness, so restoration requires the fusion of RGB appearance, event changes, and a definition of the desired target exposure.
Previous event-guided low-light studies, including EvLowLight \cite{liang2023coherent} and EvLight \cite{chen2025evlight++}, have provided experimental evidence that event measurements can complement RGB inputs in difficult illumination.
SEE-Net \cite{lu2025SEE} subsequently extended this setting from fixed low-to-normal enhancement to brightness adjustment across low, normal, and bright exposures.
Together, these studies motivate a benchmark that evaluates event-guided restoration across both underexposed and overexposed inputs.

Broad-range brightness adjustment couples several open design choices.
A complete system must represent asynchronous events, fuse them with RGB appearance, condition the desired brightness, and decide how much temporal context to use.
A shared challenge makes these choices observable under one data and evaluation protocol and provides reference points for subsequent research.
Organizing the challenge with the Event-Based Multimodal Vision Workshop also creates a focused forum for the event-vision community to compare system designs, discuss reproducibility, and identify limitations that are difficult to see from a single baseline.

The SEE Challenge 2026 extends evaluation on SEE-600K~\cite{lu2025SEE} beyond a single baseline.
It supports a community study of independent event-guided restoration systems on newly recorded hidden sequences.
Around 70 teams registered interest, 15 accounts produced valid Phase 2 submissions, and six distinct teams completed organizer-side technical verification.
Their systems cover cross-attention, feature modulation, state-space models, latent flow, temporal averaging, cascaded restoration, and test-time augmentation.
This diversity allows the report to document both current performance and the practical trade-offs between accuracy, temporal context, model size, and inference multiplicity.

This report makes four organizer-side contributions.
First, it defines the task and documents data construction, event--frame synchronization, the target-brightness statistic, and the open-system evaluation protocol.
Second, it reports a verified ranking based on organizer-run dense evaluation of the six systems for which checkpoints, code, and instructions were provided.
Third, it analyzes results across exposure settings, motion and texture cases, and a shared failure example using pixel-level, structural, and perceptual metrics.
Finally, it compares the architectural, conditioning, temporal, and inference choices of the submitted systems and identifies priorities for future challenge editions.
Because the competition permits different temporal contexts, pretrained weights, TTA, cascades, and post-processing, the ranking evaluates complete systems and does not isolate the contribution of any single component.

\section{Challenge Protocol}
\label{sec:protocol}

\subsection{Task Definition and Open-System Scope}

For an evaluation sample, let $\mathbf{I}_{\mathrm{in}}=\{I_{\mathrm{in}}^{(k)}\}_{k=1}^{K}$ denote one or more input RGB frames and $\mathbf{E}=\{E^{(k)}\}_{k=1}^{K}$ their associated event data.
The organizers provide a scalar target-brightness statistic $B$ for the target reference.
A submitted system predicts

\begin{equation}
\hat{I}_{\mathrm{out}}=F(\mathbf{I}_{\mathrm{in}},\mathbf{E},B),
\end{equation}

and the prediction is compared with the hidden reference $I_{\mathrm{gt}}$.
The evaluated mappings are Low-to-Normal and High-to-Normal restoration.
The former brightens an underexposed input and recovers visible structure, whereas the latter reduces the brightness of an overexposed input and reconstructs clipped or weakened detail.

The challenge is an offline open-system evaluation.
No maximum value of $K$ or maximum temporal span was imposed, and participants were permitted to use both preceding and future adjacent frames from the provided sequence.
Participants could also choose their event representation, fusion strategy, training objective, and inference pipeline.
No constrained single-frame or single-pass track was defined retrospectively; temporal context and inference multiplicity are therefore disclosed as system properties rather than controlled variables.

\subsection{Dataset Splits and Pair Construction}

SEE-600K contains 610,126 image-event observations from 202 released real-world scenes~\cite{lu2025SEE}.
Each scene contains, on average, four recordings of the same environment and robotic-arm trajectory under different light-transmission, aperture, and exposure settings.
Neutral-density filters with different transmission levels create low-light, normal-light, and high-light recordings while the Universal Robots UR5e repeats the same trajectory.
After temporal registration, frames at corresponding trajectory positions are paired across exposure recordings.
An underexposed or overexposed frame can therefore be used as the source and a temporally corresponding normal-light frame as the target.
When a scene contains multiple eligible source or normal-light recordings, more than one restoration mapping can be constructed from the same underlying image-event observations.
This is why the number of constructed restoration pairs differs from the number of distinct image-event observations.

\begin{table*}[t]
\centering
\caption{Data used by the challenge. Pair counts refer to constructed source--target restoration pairs rather than distinct raw frames. The three splits are scene-disjoint.}
\label{tab:data_protocol}
\small
\setlength{\tabcolsep}{6pt}
\resizebox{\textwidth}{!}{%
\begin{tabular}{lrrlll}
\toprule
Split & Scenes & Pairs & Reference & Purpose & Motion setting \\
\midrule
Training split & 165 & 509,132 & Public & Training & Camera motion\\
Test split & 37 & 147,368 & Public & Phase 1 validation & Camera motion \\
Hidden final split & 6 & 3,405 & Hidden & Phase 2 validation & Camera and scene motion \\
\bottomrule
\end{tabular}
}
\vspace{-10pt}
\end{table*}

The 165-scene training split and 37-scene released test split together form the 202 released SEE-600K scenes and yield 656,500 constructed restoration pairs.
The released references were available to all participants and supported training and Phase 1 validation.
The final evaluation used six additional newly recorded scene groups.
Only the input RGB data, events, metadata, and target-brightness statistic were released for these groups; their normal-light references remained hidden.

\subsection{Event--Frame Synchronization and Data Format}

The data were recorded with a color DAVIS346 sensor at a spatial resolution of $346\times260$ pixels.
APS frames, events, and IMU measurements share the sensor timestamp domain.
Let $t^{(k)}_{\mathrm{exp}}$ denote the exposure-start timestamp of RGB frame $k$.
The organizer-provided event interval associated with that frame is

\begin{equation}
E^{(k)}=\{e_i=(x_i,y_i,t_i,p_i)\mid t^{(k)}_{\mathrm{exp}}\leq t_i<t^{(k+1)}_{\mathrm{exp}}\}.
\end{equation}

The half-open interval prevents an event at a boundary from being assigned twice.
The released data contain 8-bit RGB PNG frames and timestamped event arrays with coordinates and polarity; the raw recordings were originally stored in AEDAT4 format together with IMU measurements.
The challenge does not prescribe a voxelization or event-window aggregation strategy.
Participants may transform the raw events into any representation and may combine events from multiple adjacent frame intervals when using temporal context.

Repeated robotic-arm trajectories are temporally registered using the synchronized 1-kHz IMU signals~\cite{lu2025SEE}.
The dataset paper evaluates spatial alignment by matching SIFT features with FLANN, estimating an affine transformation with RANSAC, applying the transformation across the image, and averaging the resulting pixel displacement.
The reported mean displacement is 0.2967 pixels, with a temporal registration error of at most approximately 1 ms~\cite{lu2025SEE}.
These values characterize the acquisition and registration procedure.

\subsection{Target-Brightness Statistic}

For each hidden reference image $I_{\mathrm{gt}}\in[0,1]^{H\times W\times3}$, the target-brightness statistic is the arithmetic mean over all pixels and RGB channels:

\begin{equation}
B=\frac{1}{3HW}\sum_{y=1}^{H}\sum_{x=1}^{W}\sum_{c\in\{R,G,B\}}I_{\mathrm{gt}}(y,x,c).
\end{equation}

The hidden reference is read as an 8-bit sRGB image and converted to float32 by division by 255.
The mean is computed directly in the normalized sRGB domain, without inverse-gamma conversion, luminance weighting, a median operator, or separate per-channel prompts.
Thus, $B\in[0,1]$ is a single float32 value computed independently for every target frame.

In the official evaluation, $B$ is calculated by the organizers from the hidden reference and supplied with the test input.
Participants cannot access $I_{\mathrm{gt}}$, but they can use or ignore the scalar derived from it.
Accordingly, $B$ is privileged ground-truth-derived information, and the official benchmark is more precisely described as target-brightness-statistic-conditioned restoration.
Although an individual model may technically accept a user-selected value at inference time, the challenge results do not evaluate generalization to arbitrary user-selected brightness values.

The target-brightness statistic is needed because brightness adjustment can be a one-to-many mapping before the target is specified.
Let $X=(\mathbf{I}_{\mathrm{in}},\mathbf{E})$ denote the RGB-event input.
For the same input $X$, different target brightness levels may correspond to different valid reference images.
Without the brightness statistic, the optimal prediction under a squared reconstruction loss is

\begin{equation}
G^{*}(X)
=
\arg\min_{\hat{I}}
\mathbb{E}
\left[
\left\|\hat{I}-I_{\mathrm{gt}}\right\|_{2}^{2}
\mid X
\right]
=
\mathbb{E}
\left[
I_{\mathrm{gt}}\mid X
\right].
\end{equation}

Therefore, if the training data contain several target brightness levels for the same input, the prediction may approach their conditional average.
Its brightness is then determined by the conditional target distribution rather than a specified target level.
By conditioning the model on $B$, the optimal prediction becomes

\begin{equation}
F^{*}(X,B)
=
\arg\min_{\hat{I}}
\mathbb{E}
\left[
\left\|\hat{I}-I_{\mathrm{gt}}\right\|_{2}^{2}
\mid X,B
\right]
=
\mathbb{E}
\left[
I_{\mathrm{gt}}\mid X,B
\right].
\end{equation}

Conditioning on $B$ therefore reduces target ambiguity by selecting the desired brightness level.
This derivation explains the role of $B$ under the stated loss.

\subsection{Evaluation Protocol and Challenge Rules}

Predictions are evaluated in normalized RGB space without boundary cropping.
PSNR is the primary metric and determines the verified ranking; SSIM is secondary.
Both metrics are computed per frame and averaged separately over the Low-to-Normal and High-to-Normal subsets, and the reported average is the mean of the two subset scores.
L1 and LPIPS are additionally reported for the organizer-selected case analysis, but they do not affect the official ranking.
Parameters and FLOPs are descriptive and do not affect the ranking.

Because of CodaBench storage and evaluation limits, the Phase 2 server evaluates one frame in every ten from each hidden video.
For technical verification, the organizers run the submitted checkpoints and inference code on the same hidden videos using dense temporal sampling.
All accounts were subject to the same phase-specific server settings and were limited to five submissions per day and 100 submissions per account in each phase.
The challenge website opened on May 10, 2026; the validation server opened on May 25; the hidden test data and Phase 2 server opened on June 25; submissions closed on July 3; and results were announced on July 10.
External public data and pretrained weights were permitted, provided that their use was disclosed in the technical report.
As noted above, the rules did not restrict temporal context and allowed future frames in this offline setting.

A valid CodaBench submission is a correctly formatted prediction package that was successfully evaluated by the Phase 2 server.
Fifteen such submissions/accounts were received.
Because a platform account does not by itself establish a distinct verified team identity, these submissions are not interpreted as 15 verified teams.
The six teams reported in this paper responded to the organizer invitation, confirmed their team identities, described their systems, and provided checkpoints, inference code, and instructions for organizer-side evaluation.
Only these six systems form the verified open-system ranking in this report.

\subsection{Prior Evidence and Official Baseline}

The effectiveness of event guidance is an established premise of the challenge rather than a conclusion drawn from its open-system ranking.
Prior event-guided low-light studies reported controlled comparisons supporting the complementary value of events under degraded illumination~\cite{liang2023coherent,liu2023low,chen2025evlight++}.
SEE-Net~\cite{lu2025SEE} extended this line of work to broad-range brightness adjustment and analyzed brightness conditioning.
Its previously reported prompt-merge ablation on SDE reduced PSNR from 23.57 dB to 22.26 dB when prompt merging was disabled, while SSIM changed from 0.7724 to 0.7713; it also visualized outputs over a sweep of prompt values.
These are results from the prior SEE-Net study.

The released SEE-Net baseline takes one RGB frame, a synchronized 64-bin event voxel grid, and a brightness statistic.
It uses separate RGB and event encoders, event-aware cross-attention, and a five-layer exposure decoder.
The model has 1.89M parameters and obtains 18.8279 dB PSNR and 0.6414 SSIM on the released test split.
These released-split results are provided for reference.

\section{Challenge Results and Organizer-Side Analysis}
\label{sec:results}

\begin{table*}[t]
    \centering
\caption{
Results of the SEE Challenge 2026 under Phase 2 Codabench and organizer-side dense sampling.
Dense-sampling PSNR determines the ranking, while SSIM is secondary.
Best and second-best results are shown in \textbf{bold} and \underline{underlined}.
FLOPs (G) and parameters are reported at the native resolution.
FLOPs (G) include cascaded stages but exclude TTA, temporal averaging, ensembles, and post-processing.
}
    \label{tab:challenge_results}
    \small
    \setlength{\tabcolsep}{5.2pt}
    \renewcommand{\arraystretch}{1.15}
    \resizebox{\textwidth}{!}{
    \begin{tabular}{@{}lccccccrr@{}}
        \toprule
        \multicolumn{9}{l}{\textbf{(a) Phase 2 evaluation on Codabench}} \\[2pt]
        \cmidrule(lr){1-9}
        \multirow{2}{*}{Participant}
        & \multicolumn{2}{c}{Average}
        & \multicolumn{2}{c}{Low-to-Normal}
        & \multicolumn{2}{c}{High-to-Normal}
        & \multirow{2}{*}{FLOPs (G)}
        & \multirow{2}{*}{Params (M)} \\
        \cmidrule(lr){2-3}
        \cmidrule(lr){4-5}
        \cmidrule(lr){6-7}
        & PSNR & SSIM
        & PSNR & SSIM
        & PSNR & SSIM
        & & \\
        \midrule
        pixelartai
        & \textbf{23.1810} & \textbf{0.7246}
        & \textbf{21.2066} & \textbf{0.6965}
        & \textbf{25.1555} & \textbf{0.7527}
        & 1314.77 & 10.24 \\
        vincent2013
        & \underline{23.0865} & \underline{0.7178}
        & 21.1040 & \underline{0.6941}
        & \underline{25.0689} & \underline{0.7416}
        & 3390.56 & 54.21 \\
        EventFuse
        & 22.7992 & 0.7057
        & \underline{21.1942} & 0.6809
        & 24.4042 & 0.7305
        & 97.51 & 1.16 \\
        bucloud
        & 22.6168 & 0.7048
        & 20.7068 & 0.6779
        & 24.5268 & 0.7316
        & 870.37 & 15.85 \\
        bin\_jiang
        & 22.1735 & 0.6853
        & 20.6521 & 0.6632
        & 23.6948 & 0.7074
        & 346.07 & 3.06 \\
        aka18
        & 20.8262 & 0.6608
        & 19.2888 & 0.6452
        & 22.3636 & 0.6764
        & 8141.41 & 3.38 \\
        \midrule
        \midrule
        \addlinespace[3pt]
        \multicolumn{9}{l}{
        \textbf{(b) Official dense-sampling evaluation}
        } \\[2pt]
        \cmidrule(lr){1-9}
        \multirow{2}{*}{Participant}
        & \multicolumn{2}{c}{Average}
        & \multicolumn{2}{c}{Low-to-Normal}
        & \multicolumn{2}{c}{High-to-Normal}
        & \multirow{2}{*}{FLOPs (G)}
        & \multirow{2}{*}{Params (M)} \\
        \cmidrule(lr){2-3}
        \cmidrule(lr){4-5}
        \cmidrule(lr){6-7}
        & PSNR & SSIM
        & PSNR & SSIM
        & PSNR & SSIM
        & & \\
        \midrule
        pixelartai
        & \textbf{23.1731} & \textbf{0.7237}
        & \underline{21.1863} & \textbf{0.6957}
        & \textbf{25.1600} & \textbf{0.7518}
        & 1314.77 & 10.24 \\
        vincent2013
        & \underline{23.0873} & \underline{0.7169}
        & 21.1212 & \underline{0.6942}
        & \underline{25.0534} & \underline{0.7396}
        & 3390.56 & 54.21 \\
        EventFuse
        & 22.8043 & 0.7061
        & \textbf{21.1974} & 0.6810
        & 24.4112 & 0.7311
        & 97.51 & 1.16 \\
        bucloud
        & 22.5865 & 0.7038
        & 20.7150 & 0.6780
        & 24.4581 & 0.7297
        & 870.37 & 15.85 \\
        bin\_jiang
        & 22.1478 & 0.6849
        & 20.6705 & 0.6637
        & 23.6251 & 0.7061
        & 346.07 & 3.06 \\
        aka18
        & 20.8082 & 0.6606
        & 19.3104 & 0.6453
        & 22.3061 & 0.6758
        & 8141.41 & 3.38 \\
        \bottomrule
    \end{tabular}
    }
\vspace{-15pt}
\end{table*}

\noindent\textbf{Verified Open-System Ranking.}
Table~\ref{tab:challenge_results} reports the six verified systems under both Phase 2 CodaBench sampling and organizer-side dense sampling.
MLSLabs (CodaBench handle: pixelartai) ranks first under dense sampling with 23.1731 dB PSNR and 0.7237 SSIM, followed by ACVLab-TL (vincent2013) with 23.0873 dB and 0.7169 SSIM.
The PSNR difference between the top two systems is only 0.0858 dB.
We therefore describe their aggregate scores as closely spaced and do not interpret the average difference as evidence of a statistically meaningful separation.

Dense sampling preserves the ordering observed on CodaBench.
Across all six systems, the absolute change from subsampled to dense evaluation is below 0.04 dB PSNR and 0.001 SSIM.
This agreement indicates that the one-in-ten CodaBench sampling provides a close estimate of the dense-sequence averages for these submissions, while the organizer-side run additionally verifies that the supplied code and checkpoints reproduce the submitted predictions.

\noindent\textbf{Performance Across Exposure Settings.}
All six systems obtain lower PSNR on Low-to-Normal than on High-to-Normal restoration.
For pixelartai, for example, the dense scores are 21.1863 dB and 25.1600 dB, respectively.
The consistent gap shows that the Low-to-Normal subset is empirically more difficult under this hidden-test composition.
It is consistent with the hypothesis that severe underexposure provides weaker and noisier RGB measurements, but the aggregate scores alone do not establish that mechanism; scene content, motion, event density, target distribution, and residual registration error may also contribute.

\begin{table*}[t!]
\centering
\caption{Performance comparison across different groups.}
\label{tab:group_results}
\resizebox{\textwidth}{!}{
\begin{tabular}{l|cccc|cccc|cccc}
\toprule
& \multicolumn{4}{c|}{Robot Motion (G1)}
& \multicolumn{4}{c|}{Checkerboard (G2)}
& \multicolumn{4}{c}{Chinese Text (G3)} \\
Team
& PSNR$\uparrow$ & SSIM$\uparrow$ & L1$\downarrow$ & LPIPS$\downarrow$
& PSNR$\uparrow$ & SSIM$\uparrow$ & L1$\downarrow$ & LPIPS$\downarrow$
& PSNR$\uparrow$ & SSIM$\uparrow$ & L1$\downarrow$ & LPIPS$\downarrow$ \\
\midrule
pixelartai
& 18.76 & 0.4998 & 0.0826 & 0.2767
& 21.49 & 0.7951 & 0.0638 & 0.1553
& 24.83 & 0.7486 & 0.0375 & 0.1838 \\
vincent2013
& 18.85 & 0.5010 & 0.0805 & 0.2701
& 21.32 & 0.7941 & 0.0657 & 0.1548
& 25.06 & 0.7583 & 0.0364 & 0.1800 \\
EventFuse
& 18.73 & 0.4889 & 0.0847 & 0.3100
& 21.80 & 0.7902 & 0.0622 & 0.1556
& 24.40 & 0.7498 & 0.0427 & 0.1877 \\
bucloud
& 18.71 & 0.4956 & 0.0848 & 0.3069
& 21.17 & 0.7814 & 0.0668 & 0.1694
& 24.52 & 0.7484 & 0.0400 & 0.1777 \\
jiangbin
& 18.79 & 0.4986 & 0.0852 & 0.3431
& 21.11 & 0.7603 & 0.0706 & 0.1748
& 23.69 & 0.7299 & 0.0467 & 0.2217 \\
aka18
& 17.07 & 0.4735 & 0.1122 & 0.3772
& 20.43 & 0.7513 & 0.0749 & 0.1996
& 22.36 & 0.6954 & 0.0557 & 0.2107 \\
SEE-Net
& 14.81 & 0.4482 & 0.1552 & 0.4610
& 16.87 & 0.7037 & 0.1148 & 0.2564
& 21.15 & 0.6309 & 0.0695 & 0.2560 \\
\bottomrule
\end{tabular}}
\end{table*}

\noindent\textbf{Representative Case Analysis.}
Table~\ref{tab:group_results} reports three representative hidden groups selected to probe robot motion, repetitive checkerboard texture, and Chinese text.
The cases provide interpretable comparisons.
Among the three cases, Robot Motion (G1) yields the lowest PSNR for every submitted system.
ACVLab-TL (vincent2013) leads this case on PSNR, SSIM, L1, and LPIPS, with 18.85 dB PSNR and 0.5010 SSIM.
The uniformly lower scores are consistent with motion being an additional challenge, although one group cannot isolate motion from illumination, scene content, event density, or registration effects.

Checkerboard (G2) tests repetitive high-frequency structure.
EventFuse obtains the best PSNR and L1, MLSLabs (pixelartai) obtains the best SSIM, and ACVLab-TL obtains the best LPIPS.
The disagreement is small but informative: the three metrics favor different reconstructions of the same fine texture.
Chinese Text (G3) produces the highest PSNR of the three cases for all submitted systems.
ACVLab-TL leads PSNR, SSIM, and L1, whereas ACVLab-LCY (bucloud) obtains the lowest LPIPS.
This split again shows that pixel fidelity, structural similarity, and perceptual distance need not select the same system.
All submitted systems outperform the released SEE-Net baseline in the three cases, but the gains and metric ordering depend on the visual content.
Consequently, the average leaderboard should be read together with content-specific evidence.

\noindent\textbf{Organizer-Side Synthesis.}
Taken together, the results support three observations.
First, dense sampling confirms the aggregate ordering obtained by the server, but the 0.0858 dB gap between the top two systems remains too small to support a claim of clear separation.
Second, the named cases expose content-dependent behavior that is hidden by the average: motion lowers all reported PSNR values, repetitive texture produces metric disagreement, and Chinese text changes the relative perceptual ranking.
Third, the shared failure case shows that visibility recovery and detail recovery are distinct objectives under severe underexposure.
The disclosed system configurations help formulate hypotheses about temporal context, TTA, and model scale, but the open-system ranking cannot attribute any observation to one design choice.
A future challenge should therefore retain an open track while adding predeclared content strata and a constrained single-frame track for controlled comparisons.

\noindent\textbf{System Design and Inference Cost.}
The verified systems differ in backbone scale, event fusion, pretraining, temporal processing, and inference multiplicity.
Their parameter counts range from 1.16M for EventFuse to 54.21M for the two-stage EventRestormer pipeline, but model size alone does not explain the ranking.
Only Event-SCAM uses adjacent frames: its core network processes one RGB--event pair per call and conditionally averages outputs from the target frame and six neighboring frames.
The other five systems receive a single target frame and its associated events, even when they repeat inference on that input.
EventFuse performs one compact restoration pass; EventRestormer combines a two-stage cascade with eight-way TTA, resulting in 16 stage-level forward passes per output.
EvLCD and SEE-SplitNet each use four single-frame TTA variants, while REFM performs six Euler updates within one single-frame latent-transport pipeline.

The FLOPs in Table~\ref{tab:challenge_results} describe the reported core networks and are not a standardized end-to-end cost measurement.
Moreover, the teams reported latency on different GPUs and software environments.
We therefore use the disclosed frame counts and inference passes to interpret efficiency and avoid presenting the reported FLOPs or latency as a controlled hardware comparison.

\begin{figure*}[t!]
\centering
\includegraphics[width=1\textwidth]{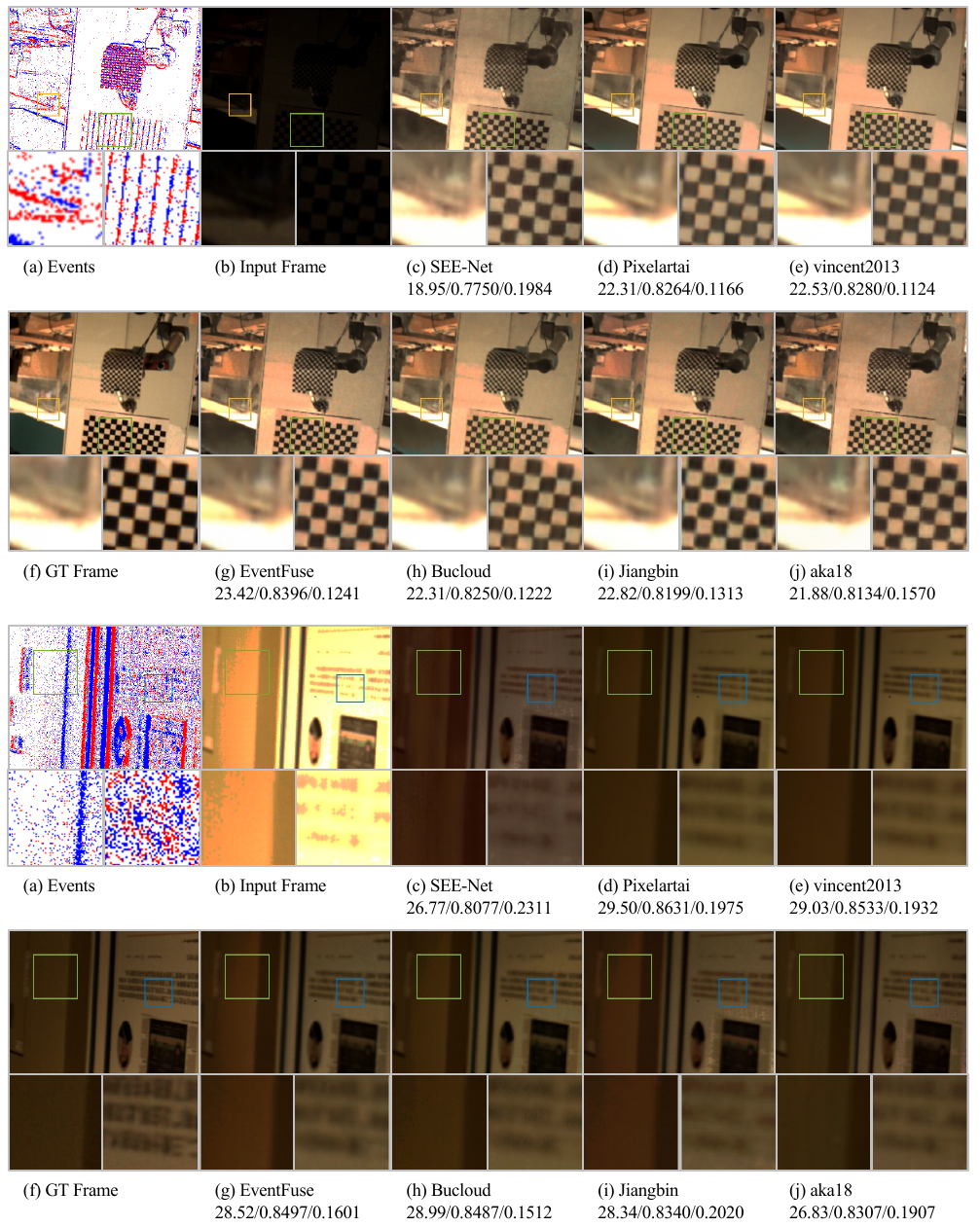}
\caption{\small Qualitative comparison of the verified submissions on representative Low-to-Normal and High-to-Normal examples. Input RGB frames, event visualizations, reference images, restored outputs, and enlarged difficult regions are displayed with a common image range and without additional gamma or contrast adjustment. The examples illustrate differences in residual noise, color, local contrast, and texture recovery that are not fully captured by average PSNR and SSIM.}
\label{fig:final_leaderboard}
\end{figure*}

\noindent\textbf{Qualitative Observations.}
Figure~\ref{fig:final_leaderboard} complements the numerical results.
The verified systems generally recover substantially more visible structure than the released baseline, especially in strongly degraded regions.
However, difficult examples still exhibit residual noise, color shifts, weakened texture, or local contrast errors.
The event visualization helps identify regions containing temporal contrast information, but its presence does not guarantee recovery when the RGB measurement is severely clipped or when the event signal is sparse or noisy.
These examples motivate evaluating both pixel-level fidelity and perceptual quality and show that the current task remains unsolved despite the improvement over the baseline.

\clearpage
\begin{figure*}[t!]
\centering
\includegraphics[width=1\textwidth]{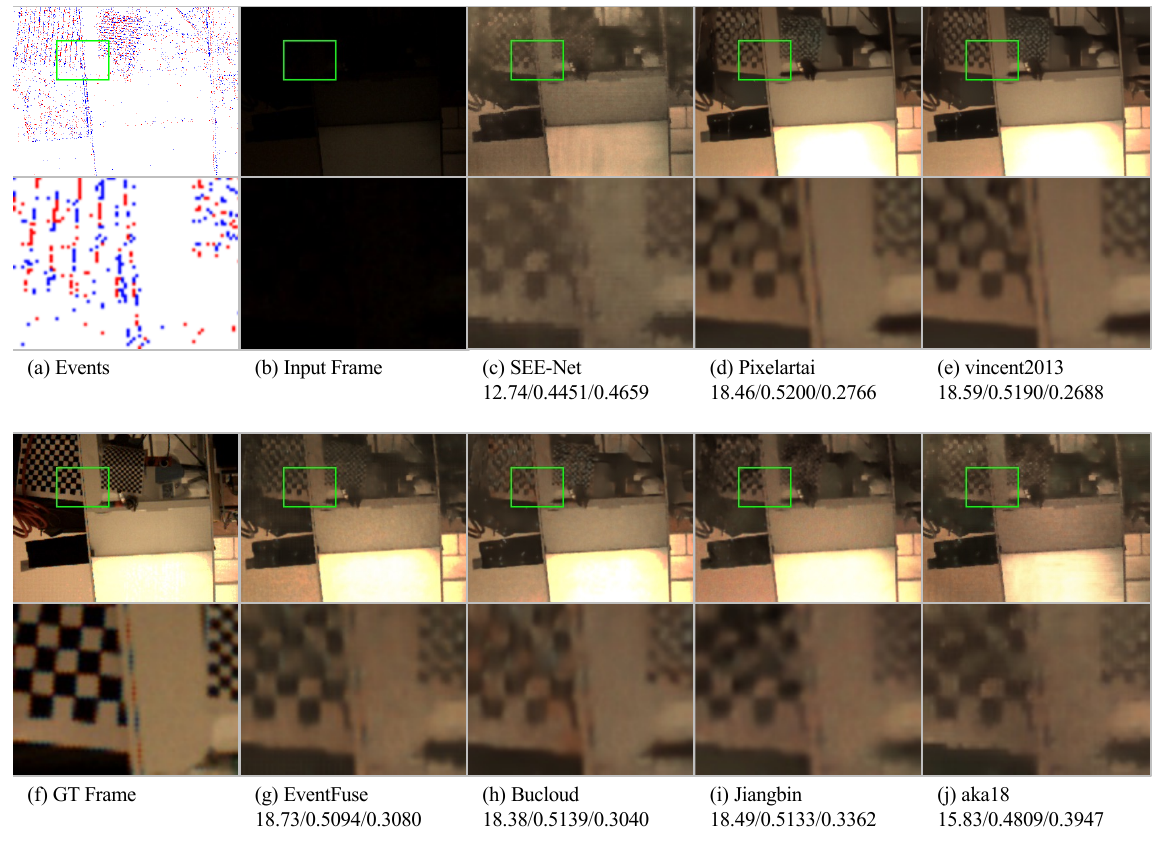}
\caption{\small Representative shared failure case under severe underexposure. (a) Accumulated events, (b) input frame, (c) SEE-Net baseline, (d)--(e) and (g)--(j) verified submissions, and (f) ground-truth frame. The green boxes indicate the enlarged regions shown below. Although the submitted systems substantially brighten the input, none faithfully reconstructs the checkerboard structure. The values below each result denote PSNR/SSIM/LPIPS ($\uparrow/\uparrow/\downarrow$).}
\label{fig:failure_cases}
\end{figure*}

\noindent\textbf{Shared Failure Pattern.}
Figure~\ref{fig:failure_cases} presents a severely underexposed example for which none of the evaluated systems faithfully reconstructs the local checkerboard structure.
All verified submissions improve substantially over the SEE-Net baseline, which obtains 12.74 dB PSNR, 0.4451 SSIM, and 0.4659 LPIPS.
Nevertheless, the best submitted PSNR is only 18.73 dB, the best SSIM is 0.5200, and the best LPIPS is 0.2688; these values are achieved by different systems.
The methods largely correct global brightness, but their enlarged crops share blurred square boundaries, weakened contrast, and color shifts relative to the reference.
The event visualization contains activity around the scene structure, yet the nearly black RGB crop provides weak appearance and color measurements.
This single example does not isolate the cause of failure, but it identifies a common limitation: event-guided exposure correction can restore visibility without recovering fine local texture faithfully.

\section{Participating Methods}
\label{sec:methods}

This section summarizes the six verified systems using a common set of dimensions.
Complete architecture diagrams, training configurations, and team-provided descriptions are included in the supplementary material.
Table~\ref{tab:method_strategies} provides a standardized comparison of pretraining, brightness conditioning, TTA, temporal processing, and inference multiplicity.
It separates multi-frame temporal processing from repeated inference on one frame: Event-SCAM is the only submitted system that aggregates adjacent-frame outputs, while the other five systems use a single target-frame input.
The descriptions below focus on the design choices needed to interpret the challenge results.

\begin{table*}[t!]
\centering
\caption{
Comparison of training and inference strategies adopted by the participating teams and the SEE baseline.
All methods use the brightness prompt.
\underline{Effective Passes} denotes the number of model forward passes or iterative updates used during inference.
}
\label{tab:method_strategies}
\setlength{\tabcolsep}{10pt}
\resizebox{\textwidth}{!}{
\begin{tabular}{lccccc}
\toprule
Team
& Pretraining
& Prompt
& TTA
& Temporal Processing
& Effective Passes \\
\midrule
pixelartai   & $\times$     & $\checkmark$ & $\times$     & $\checkmark$ & 1 \\
vincent2013  & $\checkmark$ & $\checkmark$ & $\checkmark$ & $\times$      & 16 \\
EventFuse    & $\times$     & $\checkmark$ & $\times$     & $\times$      & 1 \\
bucloud      & $\times$     & $\checkmark$ & $\checkmark$ & $\times$      & 4 \\
jiangbin     & $\checkmark$ & $\checkmark$ & $\times$     & $\times$      & 1 \\
aka18        & $\times$     & $\checkmark$ & $\checkmark$ & $\times$      & 4 \\
\midrule
SEE-Net          & $\times$     & $\checkmark$ & $\times$     & $\times$     & 1 \\
\bottomrule
\end{tabular}
}
\end{table*}

\noindent\textbf{MLSLabs (pixelartai).}
Event-SCAM uses NAFBlocks and bidirectional cross-attention to fuse RGB and event features.
It is trained for 300 epochs on SEE-600K with Charbonnier and gradient losses, without external data or pretrained weights.
The core network processes one frame--event pair per call.
When the event count is below 3200, the method averages the target-frame output with three preceding and three succeeding outputs; otherwise, it returns only the target-frame prediction.
Event-SCAM is the only multi-frame submission and uses up to seven single-frame calls.

\noindent\textbf{ACVLab-TL (vincent2013).}
EventRestormer uses a four-level Restormer RGB backbone, a 64-bin event encoder, and spatial FiLM fusion.
A seven-dimensional vector formed from $B$ and RGB statistics conditions its latent and decoder stages, while public Restormer denoising weights initialize the RGB backbone.
Its two-stage cascade applies eight-way dihedral TTA to the same target frame and event grid, producing 16 stage-level passes without adjacent frames.

\noindent\textbf{EventFuse.}
EventFuse processes one RGB frame and a 64-bin event voxel grid using Bayer and coordinate encodings, a Mamba-based illumination estimator, bidirectional fusion, and sparse state-space blocks.
Its decoder is conditioned on $B$, and the model is trained on the official data without external data or pretrained weights.
With 1.16M parameters, it is the smallest verified model and restores each frame in one pass.

\noindent\textbf{ACVLab-LCY (bucloud).}
EvLCD estimates an exposure-corrected base image from local color distributions and predicts a residual with a 64-bin event encoder and FiLM-conditioned U-Net.
It is trained from scratch on the official data in two stages.
Inference averages four flipped predictions of the same frame and shifts their global mean to match $B$.

\noindent\textbf{VisionLab-JB (bin\_jiang).}
REFM treats restoration from one RGB--event input as rectified-flow transport in a learned RGB latent space.
A velocity field fuses a binary event voxel grid with an exposure descriptor derived from source brightness and $B$, then performs six Euler updates from the observed latent state.
The 3.06M-parameter pipeline uses only the official training data and decodes the updated latent representation once.

\noindent\textbf{aka18.}
SEE-SplitNet trains wider exposure-specific SEE-Net variants for Low-to-Normal and High-to-Normal restoration.
It sets $C_1=128$ and $C_2=128$, conditions on $B$ with a 7-layer exposure MLP, and uses progressive patch-size training followed by PSNR-oriented fine-tuning.
Inference selects the task-specific model and applies four TTA variants to the same target frame.

Event fusion differs across the submitted architectures.
Five systems use one target frame.
Event-SCAM instead conditionally aggregates adjacent-frame outputs.
Repeated inference and multi-frame processing are separate system properties.

\begin{table}[t!]
\centering
\captionsetup{hypcap=false}
\captionof{table}{Contributing teams, participant names, and affiliations for the verified systems summarized in this paper.}
\label{tab:participants}
\scriptsize
\setlength{\tabcolsep}{5pt}
\renewcommand{\arraystretch}{1.15}
\begin{tabular}{@{}p{0.18\textwidth}p{0.38\textwidth}p{0.38\textwidth}@{}}
\toprule
Team & Participants & Affiliation \\
\midrule
MLSLabs (pixelartai) & Dongyang Zhang, Renjie Zou, Zhiwei Huang, Dong Jiang & Malanshan Audio \& Video Laboratory \\
\hline
ACVLab-TL (vincent2013) & Yun-Tze Tsai, Shao-Kai Liu, Chia-Ming Lee, Chih-Chung Hsu & Advanced Computer Vision Lab (ACVLab), National Yang Ming Chiao Tung University \\
\hline
EventFuse & Yixin Chen, Lupeng Liu & University of Chinese Academy of Sciences \\
\hline
ACVLab-LCY (bucloud) & Chia-Yu Lin & Advanced Computer Vision Lab (ACVLab), National Yang Ming Chiao Tung University; National Cheng Kung University \\
\hline
VisionLab-JB (bin\_jiang) & Bin Jiang & Nanjing University \\
\hline
aka18 & Tao Liu & Wuhan University \\
\bottomrule
\end{tabular}
\end{table}

\section{Limitations}
\label{sec:limitations}

The challenge and this report have several limitations.
First, the verified ranking is open-system: systems use different frame counts, future context, pretraining, cascades, TTA, temporal averaging, and post-processing.
It measures complete submitted pipelines and cannot isolate the contribution of events or any individual component.
Prior work provides the modality-level motivation, but no new matched RGB-only retraining was conducted on the hidden set.
The participating teams supplied checkpoints and inference code, so a fair organizer-run modality ablation for the submitted systems was not available.

Second, the official target-brightness statistic is derived from each hidden reference.
It is privileged information and differs from a practical setting in which a user selects a desired brightness without access to ground truth.
Although SEE-Net previously visualized different prompt values, the challenge did not conduct a standardized quantitative sweep measuring monotonicity or output-brightness error for all submitted methods.

Third, the evaluation emphasizes frame-wise PSNR and SSIM.
The representative case table adds L1 and LPIPS, and the failure figure exposes one shared qualitative weakness, but these selected examples do not replace a complete per-sequence distribution.
The challenge also does not include a temporal-consistency metric, a systematic analysis against event rate or scene motion, or a scene-level bootstrap confidence interval.
Accordingly, small score differences should not be interpreted as statistically established separations.

Finally, the reported computation and latency values were produced by teams on different hardware and may exclude parts of the full inference pipeline.
Conditional temporal averaging also makes the effective cost input-dependent.

\section{Conclusion}
We presented the SEE Challenge 2026, an open-system benchmark for event-guided, target-brightness-statistic-conditioned restoration across a broad illumination range.
Fifteen valid CodaBench submissions were received, and six distinct teams completed organizer-side verification and are included in the reported ranking.
Dense sampling produced results close to the subsampled server evaluation, while exposure-specific and content-oriented analyses showed that Low-to-Normal restoration obtains lower scores and that the best method varies across cases and metrics.
The shared failure example further shows that correcting global brightness does not ensure faithful recovery of fine local texture under severe underexposure.
The submitted systems explore diverse RGB-event fusion, conditioning, temporal, and inference strategies.
The challenge provides a common protocol and identifies open questions in exposure recovery, efficiency, temporal consistency, and practical brightness control.

\noindent\textbf{Acknowledgements:}
The organizers thank all participants and particularly the six verified teams that provided checkpoints, inference code, technical descriptions, and supplementary method reports.

\clearpage
\bibliographystyle{splncs04}
\bibliography{main}
\end{document}